\documentclass[letterpaper, 10 pt, conference]{ieeeconf}  

\IEEEoverridecommandlockouts                              

\usepackage{mathtools}
\usepackage{amsmath}
\usepackage{amsfonts}
\usepackage{amssymb}
\usepackage{dsfont}
\usepackage{algorithm}
\usepackage[noend]{algpseudocode}
\usepackage{subcaption}
\usepackage{gensymb}
\usepackage{multirow}
\usepackage{array}
\usepackage{hyperref}
\title{\LARGE \bf
Context-Aware Safe Reinforcement Learning \\ for Non-Stationary Environments
}

\author{Baiming Chen, Zuxin Liu, Jiacheng Zhu, Mengdi Xu, Wenhao Ding, Ding Zhao
\thanks{*\textit{(Corresponding author: Ding Zhao)}}
\thanks{Baiming Chen is with Tsinghua University, Beijing, China. This work was done during his visit in Carnegie Mellon University (e-mail: cbm17@mails.tsinghua.edu.cn)}%
\thanks{Zuxin Liu, Jiacheng Zhu, Mengdi Xu, Wenhao Ding and Ding Zhao are with the Department of Mechanical Engineering, Carnegie Mellon University, USA (e-mail: \{zuxinl, jzhu4, mengdixu, wenhaod, dingzhao\}@andrew.cmu.edu)}
}

\begin{document}
\bstctlcite{IEEEexample:BSTcontrol}

\maketitle
\thispagestyle{empty}
\pagestyle{empty}

\begin{abstract}

Safety is a critical concern when deploying reinforcement learning agents for realistic tasks. Recently, safe reinforcement learning algorithms have been developed to optimize the agent's performance while avoiding violations of safety constraints. 
However, few studies have addressed the non-stationary disturbances in the environments, which may cause catastrophic outcomes. In this paper, we propose the context-aware safe reinforcement learning (CASRL) method, a meta-learning framework to realize safe adaptation in non-stationary environments. We use a probabilistic latent variable model to achieve fast inference of the posterior environment transition distribution given the context data. Safety constraints are then evaluated with uncertainty-aware trajectory sampling.
The high cost of safety violations leads to the rareness of unsafe records in the dataset. We address this issue by enabling prioritized sampling during model training and formulating prior safety constraints with domain knowledge during constrained planning. The algorithm is evaluated in realistic safety-critical environments with non-stationary disturbances. Results show that the proposed algorithm significantly outperforms existing baselines in terms of safety and robustness.

\end{abstract}

\section{Introduction}

Reinforcement learning (RL) is a promising way to solve sequential decision-making tasks. 
For example, RL has shown superhuman performance in competitive games like Go~\cite{silver2016mastering} and Starcraft~\cite{vinyals2019alphastar}.
RL has also been used for the control of complex robotic systems~\cite{nagabandi2018deep,chen2020delay} such as legged robots~\cite{hwangbo2019learning}. 
However, most well-known RL algorithms~\cite{lillicrap2015continuous,schulman2017proximal,chua2018deep} do not consider safety constraints during exploration. Moreover, they are usually not adaptive to non-stationary disturbances, which are common in many realistic safety-critical applications~\cite{tobin2017domain}. These two weaknesses of current RL algorithms need to be addressed before their deployment in safety-critical environments.

\begin{figure}[t]
\centering
\includegraphics[width=0.8\columnwidth]{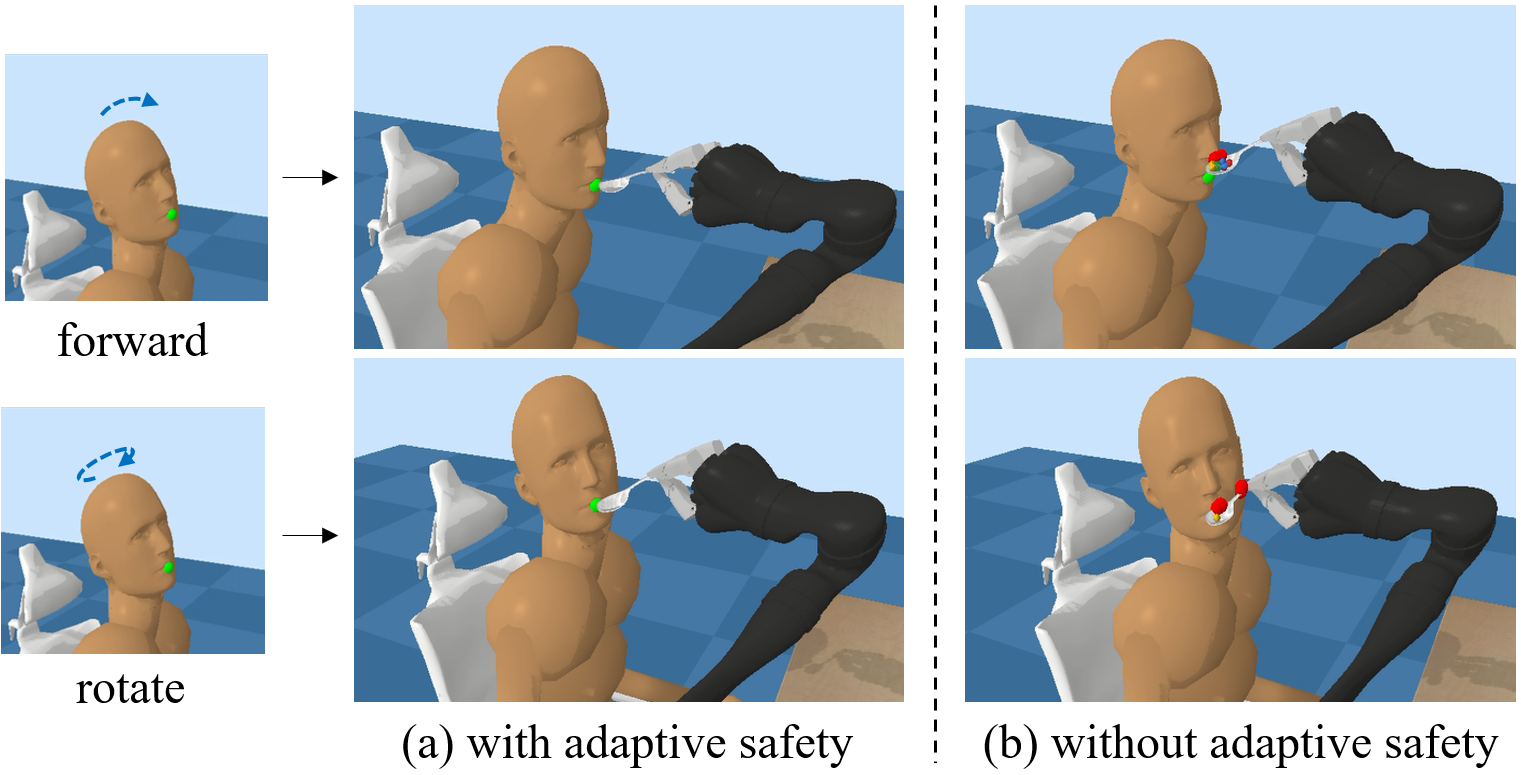} 
\caption{Healthcare environment with and without adaptive safety. Red dots indicate direct contacts between the robot and the patient which should be avoided.}
\label{fig:health_robot}
\end{figure}

Several recent studies have been proposed to address the lack of \textbf{safety} \cite{achiam2017constrained,pham2018optlayer,chow2018lyapunov,dalal2018safe} and the lack of \textbf{adaptability} \cite{finn2017model,nagabandi2018learning,xu2020task} issues of RL algorithms, respectively. However, the above two issues are entangled in realistic applications, because the environment disturbances may change the system dynamics and affect the region of safety. In other words, disturbances may cause unexpected safety violations if not properly handled. A typical example is shown in Fig.~\ref{fig:health_robot}, where a healthcare robot is trying to deliver the medicine (or food) to the patient while avoiding any direct contact.
The disturbance in this environment mainly comes from the patient's movements.
To safely finish the delivery, the robot must be able to quickly identify the patient's moving preference and adaptively generate safe control decisions. 
To the best of our knowledge, there hasn't been a general framework or a complete algorithm to fully address this entangled problem.
In this paper, we propose the context-aware safe reinforcement learning (CASRL) framework to realize safe adaptation in non-stationary environments and resolve the above entangled problem. 
Our major contribution is threefold:
\begin{enumerate}
    \item \textbf{Fast adaptation}. We study this problem under the model-based RL framework for sample efficiency. Unlike previous models that predict the next state only based on the current state and action, we use a context-aware latent variable model to infer the disturbance of the non-stationary environment based on the historical transition data, allowing task-agnostic adaptation. 
    \item \textbf{Risk-averse control}. 
    We achieve risk-averse decision making with constrained model predictive control. Constraints are used for guarantees of safety in uncertain environments. To improve exploration safety in the early stage of training, we incorporate domain knowledge to make conservative decisions with prior models. We also enable prioritized sampling of rare unsafe data during the model training to alleviate the data imbalance problem in safety-critical environments. Combined with a context-aware probabilistic model, this control regime can realize safe adaptation in non-stationary environments and resolve the aforementioned entangled problem.
    \item \textbf{Extensive evaluation}. 
    We conduct experiments in a toy example and a realistic high-dimensional environment with non-stationary disturbances. 
    Results show that the proposed method can (i) realize fast adaptation for safe control in unseen environments, (ii) scale to high-dimensional tasks, and (iii) outperform existing approaches in terms of safety and robustness.
\end{enumerate}

\section{Related Work}

\textbf{Safe reinforcement learning} has attracted long-term interest in the RL community~\cite{garcia2015comprehensive}. The Constrained Markov Decision Processes (CMDPs)~\cite{altman1999constrained} is often used to model the safe RL problem, where the agent aims to maximize its cumulative reward while satisfying certain safety constraints. Several approaches, such as the Lagrangian method~\cite{altman1998constrained} and constrained policy optimization~\cite{achiam2017constrained,chow2019lyapunov}, have been proposed to solve CMDPs. Gaussian Processes (GPs) have also been used to approximate the dynamics of the environment for safe exploration~\cite{koller2018learning,hewing2019cautious}. Particularly, Wachi and Sui~\cite{wachi2020safe} discussed the situation where the safety boundary is unknown. However, most existing safe RL methods assume a consistent environment and cannot deal with time-varying disturbances. In contrast, our method aims to realize safe control in non-stationary environments, which is more realistic for safety-critical applications. 

\begin{figure*}[t]
\centering
\includegraphics[width=\textwidth]{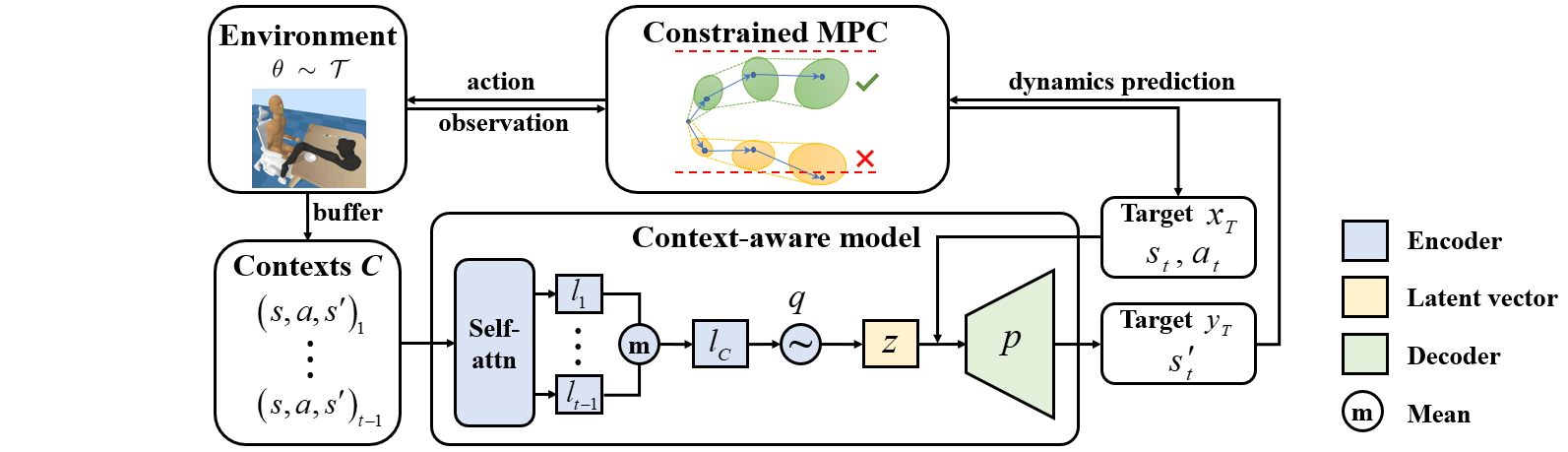} 
\caption{The flow of the proposed context-aware safe reinforcement learning (CASRL) framework. A context-aware model is used to perform conditional dynamics predictions based on the context data.}
\label{fig:npmodel}
\end{figure*}

\textbf{Robust adversarial learning} addresses the environment disturbance problem by formulating a two-player zero-sum game between the agent and the disturbance~\cite{nilim2003robust,pinto2017robust,ding2020learning}. However, the robust policies trained in this way may overfit to the worst-case scenario, so the performance is not guaranteed in other cases~\cite{rice2020overfitting}. 


\textbf{Meta-learning for RL} has recently been developed to realize adaptive control in non-stationary environments~\cite{duan2016rl,finn2017model,saemundsson2018meta,nagabandi2018learning,xu2020task,cheung2019non}. Since unsafe data are particularly rare in safety-critical environments, we focus on model-based methods for sample efficiency~\cite{chua2018deep}. Sæmundsson et al.~\cite{saemundsson2018meta} proposed to use Gaussian Processes to represent dynamics models, which may suffer from poor scalability as the dimension and the amount of data increases. Nagabandi et al.~\cite{nagabandi2018learning} integrated model-agnostic meta-learning (MAML)~\cite{finn2017model} with model-based RL. The dynamics model is represented by a neural network that uses a meta-learned initialization and is quickly updated with the latest data batch. 
However, the uncertainty is not estimated by the model, and we show that this may degrade the performance. 
Later studies from Xu et al.~\cite{xu2020task} and Nagabandi et al.~\cite{nagabandi2018deep} achieved online continual learning with streaming data by maintaining a mixture of meta-trained dynamics models. These approaches may suffer from the model explosion in complex environments where the potential number of dynamics type is large. We overcome this issue by constructing a probabilistic latent variable model that learns a continuous mapping from the disturbance space to the latent space.



\textbf{Neural Processes (NPs)}~\cite{garnelo2018neural} have been proposed for few-shot regression by learning to map a context set of input-output observations to a distribution of regression functions. Comparing to the Gaussian processes, NPs have the advantage of efficient data-fitting with linear complexity in the size of context pairs and can learn conditional distributions with a latent space. 
A later study~\cite{kim2019attentive} proposed Attentive Neural Processes (ANPs) by incorporating attention into NPs to alleviate the underfitting problem and improve the regression performance. NP-based models have shown great performance in function regression~\cite{qin2019recurrent}, image reconstruction~\cite{kim2019attentive}, and point-cloud modeling~\cite{gordon2019convolutional}. As probabilistic latent variable models, ANPs naturally enable continual online learning in continuously parameterized environments. In this paper, we will show how to incorporate ANPs for dynamics prediction and safety constraint estimation.

The rest of the paper is organized as follows. In Sec.~\ref{sec:3}, we formulate the safety-critical problem that we aim to solve in this paper. In Sec.~\ref{sec:4}, we show the inference process of unknown environment disturbances with a latent variable model. In Sec.~\ref{sec:5}, we show how to perform safe adaptation with a sampling-based model-predictive controller. The experiment results and discussions are presented in Sec.~\ref{sec:6}.


\section{Problem Statement}
\label{sec:3}
We consider non-stationary Markov Decision Processes (MDPs) with safe constraints. An MDP is defined as a tuple $(\mathcal{S}, \mathcal{A}, f, r, \gamma, \rho_0)$ where $\mathcal{S}$ denotes the state space, $\mathcal{A}$ denotes the action space, $f(s^\prime|s,a)$ is the transition distribution of the environment dynamics that takes into the current state $s\in\mathcal{S}$ and action $a\in\mathcal{A}$, and outputs the distribution of the next state $s^\prime\in\mathcal{S}$. $r(s,a)$ is the reward function, $\gamma$ is the reward discount factor, and $\rho_0$ is the distribution of the initial state. 
To simulate the disturbances in real-world environments, we consider non-stationary MDPs where the transition dynamics $f(s^\prime|s,a,\theta)$ depends on certain hidden parameters $\theta \sim \mathcal{T}$, where $\mathcal{T}$ denotes the distributions of environments parameters.
For simplicity, we assume that the environment is episodically consistent - the change of $f$ only happens at the beginning of each episode.
This setting is commonly used in related papers and can be easily generalized to other consistent time-horizons.

Denote a safe state set by $\mathcal{S}_{safe}$ and a safe action set by $\mathcal{A}_{safe}$.
The goal of safe RL is to find the optimal action sequence $a_{0:T}$ to maximize the discounted accumulated reward $\sum_{t=0}^{\tau}\gamma^t r(s_t, a_t)$, without violating the safety constraints (i.e., keeping $s_t\in\mathcal{S}_{safe}$ and $a_t\in\mathcal{A}_{safe}$ for every time step $t$). $\gamma$ is a discount factor and $\tau$ is the task horizon. Throughout this paper, we assume $\mathcal{S}_{safe}$ and $\mathcal{A}_{safe}$ are known a \textit{priori}. 


\section{Context-Aware Model Inference}
\label{sec:4}

We address the proposed problem under the model-based RL framework, where the tasks are solved by learning a dynamics model $\tilde{f}(s^\prime|s,a)$ to approximate the ground-truth environment dynamics $f(s^\prime|s,a)$. However, when the environment dynamics $f$ is non-stationary, $\tilde{f}(s^\prime|s,a)$ may fail to make accurate predictions since some hidden features of the environment are not identified. To handle this problem, we propose to learn a context-aware model $\tilde{f}(s^\prime|s,a,C)$ that performs state predictions based not only on the current state $s$ and action $a$ but also on the \textit{contexts} $C$ - the historical data collected in the current episode.
In this way, the hidden information of the environment is first inferred from $C$, and then the posterior distribution of the next state $s^\prime$ is calculated.

To incorporate domain knowledge for adaptive learning, we divide the dynamics model $\tilde{f}(s^\prime|s,a,C)$ into two parts:
\begin{subequations}
\label{equ:div}
\begin{align}
    &s^\prime \coloneqq s^\prime_h + s^\prime_g, \\
    \text{with} \quad &s^\prime_{h}\sim h(\cdot|s,a), \label{equ:div:h}  \\
    &s^\prime_{g}\sim g(\cdot|s,a,C).  \label{equ:div:g}
\end{align}
\end{subequations}


The model $h$ in Eq.~(\ref{equ:div:h}) is referred to as the \textit{prior model}. Such model can be obtained by leveraging domain knowledge without necessarily interacting with the environment, e.g., training the dynamics model in a simulator~\cite{chua2018deep} or using first principles modeling~\cite{pati2014modeling}. However, the drawback is that they are usually context-unaware.



The model $g$ in Eq.~(\ref{equ:div:g}) is called the \textit{disturbance model} (or the error model). It represents the error between the prior model $h$ and the overall dynamics model $\tilde{f}$. It is the model we aim to learn by interacting with the target non-stationary environment. The disturbance model is context-aware and should be able to capture the hidden information of the environment based on the contexts $C$. To achieve that, the disturbance model $g$ should have the following properties:
\begin{itemize}
    \item Flexibility: $g$ should be able to condition on arbitrary number of contexts to make predictions.
    \item Uncertainty awareness: $g$ should estimate the uncertainty in its predictions to balance exploration and exploitation.
    \item Scalability: $g$ should be able to scale to high-dimensional environments.
\end{itemize}

In this paper, we use an Attentive Neural Process (ANP)~\cite{kim2019attentive} to represent the disturbance dynamics model $g$ for its desirable properties and implementation simplicity. The ANP model is defined as a (infinite) family of conditional distributions, in which an arbitrary number of observed input-output \textit{contexts} $(x_C, y_C)\coloneqq(x_i,y_i)_{i\in C}$ is used to model an arbitrary number of input-output \textit{targets} $(x_T, y_T)\coloneqq(x_i,y_i)_{i\in T}$, where $C$ denotes a set of observed points and $T$ denotes a set of unobserved points (the output $y_T$ is unknown).
The ANP transforms the original conditional likelihood to a hierarchical inference structure:
\begin{equation}
    g\left(y_T|x_T, x_C, y_C\right) = \int p\left(y_T|x_T,z\right)q\left(z|l_C\right)dz
\end{equation}
where $z$ is a global latent vector describing uncertainty in the predictions of $y_T$ for given observations $(x_C, y_C)$, and is modeled by a factorized Gaussian parameterized by $l_C\coloneqq l(x_C, y_C)$, with $l$ being a deterministic function that aggregates $(x_C, y_C)$ into a fixed dimensional representation. In ANP, $l$ consists of a multilayer perceptron (MLP), self-attentions, and a mean aggregation layer to produce permutation-invariant representations.

For dynamics prediction, the input $x$ is the state-action pair $(s,a)$, and the output $y$ is the state at the next time step $s^\prime$. At time $t$, the contexts $(x_C, y_C) = (s_i,a_i,s^\prime_i)_{i\in[1:t-1]}$ contain the state-action information of the previous time steps, the target input $x_T = (s_t,a_t)$ is the current state-action pair, and we aim to predict the target output $y_T = s^\prime_t$ that represents the next state.
The flow of using context-aware model for model-based RL is shown in Fig.~\ref{fig:npmodel}. A constrained MPC controller is used for safe planning and will be introduced in the next section.

The training of ANP is based on the amortized variational inference. The parameters of the encoders and the decoder are updated by maximizing the following evidence lower bound (ELBO) with the reparametrization trick~\cite{kingma2013auto}:
\begin{equation}
\label{equ:elbo}
    \begin{split}
    &\log g\left(y_T|x_T, x_C, y_C\right) \geq\\ &\mathbb{E}_{q\left(z|l_T\right)} \left[\log g(y_T|x_T,z)\right]
    - D_{\text{KL}}\left(q\left(z|l_T\right)\parallel q\left(z|l_C\right)\right).
\end{split}
\end{equation}
where $l_T\coloneqq l(x_T, y_T)$, with $l$ being a deterministic function introduced before. The training objective of ANP can be interpreted as improving the prediction accuracy on the targets while regularizing the Kullback–Leibler divergence between the latent encoding of the contexts and the targets. 

The contexts and the targets are randomly sampled from a replay buffer that stores transition data from the same disturbance dynamics.
However, the rareness of unsafe data may lead to low prediction accuracy in the unsafe state region. To alleviate this issue, inspired by~\cite{schaul2015prioritized}, we enable prioritized experience sampling during model training - to train the context-aware model with a certain data batch, the unsafe data in this data batch are first added into the target set $T$, and then other safe data are uniformly sampled and appended to $C$ and $T$. We found that this trick can effectively increase the prediction accuracy in the unsafe region, which is discussed in Sec.~\ref{sec:pri}.

\section{Safe Adaptation with MPC}
\label{sec:5}
We formulate the safe adaptation as a constrained nonlinear optimization problem:
\begin{subequations}
\label{equ:opt}
\begin{align}
\max_{a_{0:\tau}}  \quad & \sum_{t=0}^{\tau} r(s_t, a_t) \label{equ:obj} \\
    \text{s.t.}  \quad & a_t \in \mathcal{A}_{safe} \label{equ:a_con}\\ 
    & s_{t+1} \sim \tilde{f}(\cdot|s_t, a_t, C) \label{equ:s_con1}\\
    & \Pr(s_t \notin \mathcal{S}_{safe}) \leq \delta \label{equ:s_con2}\\
    & \hat{s}_{t+1} \sim h(\cdot|\hat{s}_t, a_t) \label{equ:p_con1}\\
    & \Pr(\hat{s}_t \notin \mathcal{S}_{safe}) \leq \delta \label{equ:p_con2}\\
    \text{for} \quad & t=0,\dots,\tau \notag
\end{align}
\end{subequations}
Eq.~(\ref{equ:obj}) shows that the objective is to maximize the cumulative reward, Eq.~(\ref{equ:a_con}) represents the safety constraint on actions, and Eq.~(\ref{equ:s_con1}, \ref{equ:s_con2}) define the safety constraint on the states $s_t$ that predicted by the learned model $\tilde{f}$. Eq.~(\ref{equ:obj})-(\ref{equ:s_con2}) form the general problem of safe RL in most previous literature~\cite{koller2018learning}. However, with the non-stationary environment disturbances, the learning process of the prediction model $\tilde{f}$ may be unstable, and it is difficult for the agent to keep safe when $\tilde{f}$ is not accurate.
To alleviate this problem, we formulate the prior safety constraint shown in Eq.~(\ref{equ:p_con1}, \ref{equ:p_con2}), where a sequence of auxiliary states $\hat{s}_t$ is predicted only with the prior model $h$, and the high-probability safety constraint is applied to it ($\hat{s}_0 = s_0$). 
Though not accurate, the prior safety constraint provides extra protection for the agent based on the static prior model $h$. Applying the prior safety constraint is an effective way to incorporate domain knowledge to improve safe learning, especially when the unsafe data are expensive to obtain.
Experiment results show that it can effectively reduce the safety violation rate especially in the early stage of training (Sec.~\ref{sec:during}).

Direct solving the optimization problem Eq.~(\ref{equ:opt}) is intractable since $\tilde{f}$ is a high-dimensional nonlinear stochastic function. Previous work has used approximated uncertainty propagation techniques like sigma-point transform~\cite{ostafew2016robust} and Taylor expansion~\cite{koller2018learning} to model the state distribution as a single Gaussian distribution, and then solve Eq.~(\ref{equ:opt}) with nonlinear solvers such as the IPOPT~\cite{wachter2006implementation}. However, Deisenroth et al.~\cite{deisenroth2013gaussian} showed that the Gaussian moment matching could corrupt after long-term propagation due to the multi-modal distribution of states, inducing huge prediction errors.
Also, IPOPT cannot provide an alternative plan if no solution for Eq.~(\ref{equ:opt}) is found in limited time.

\begin{algorithm}[t]
\caption{Trajectory sampling \label{alg:traj}}
  \begin{algorithmic}
    \Procedure{TrajSampling}{$A, h, g, C, t_0$} 
    \For{SamplingTime $= 1, N$} 
    \For{$t$ = $t_0$, $t_0 + \tau_p$}
        \State $s_{th}\sim h(\cdot|s_{t-1},a_{t-1})$
        \State $s_{tg}\sim g(\cdot|s_{t-1},a_{t-1},C)$  
        \State $s_t = s_{th} + s_{tg}$ 
    \EndFor
    \EndFor
    \State \textbf{return} $\{s_{t_0:t_0 + \tau}\}_{1:N}$
    \EndProcedure
  \end{algorithmic}
\end{algorithm}

\begin{algorithm*}[t]
\caption{Context-Aware Safe Reinforcement Learning (CASRL) \label{alg:sa}}
  \begin{algorithmic}
    \State \textbf{Input:} prior model $h$, state safe set $\mathcal{X}_{safe}$, action safe set $\mathcal{A}_{safe}$, task distribution $\mathcal{T}$
    \State \textbf{Output:} disturbance model $g$, episodic replay buffer $R$
    \State $\tilde{g} \leftarrow g_0$, $R \leftarrow \{\}$ \Comment{Initialize the disturbance model and the replay buffer}
    \For{Episode = 1, $M$}
        \State $p\sim\mathcal{T}$, $C \leftarrow \{\}$, reset \text{CEM}($\cdot$), get $s_0$  \Comment{Environment sampling and episode initialization}
        \For {$t$ = 1, $\tau$}
            \For {$A\sim \text{CEM}(\cdot)$} \Comment{Sampling action sequences}
                \State $s_{t:t+\tau_p} = \textsc{TrajSampling}(A, h, g, C, s_{t-1}, t)$  \Comment{State propagation in the learned model}
                \State $\hat{s}_{t:t+\tau_p} = \textsc{TrajSampling}(A, h, 0, C,s_{t-1}, t)$  \Comment{State propagation in the prior model}
                \State $A^* = \arg \max_A \text{CVaR}_\alpha\left(\bar{R}\left(A\right)\right)$ \Comment{The optimal action sequence is selected based on the CVaR}
                \State Update CEM($\cdot$)
            \EndFor
            \State Execute $a^*_t$, get $s_{t+1}$ \Comment{$a^*_t$ is the first element of $A^*$}
            \State $C \leftarrow C \cup (s_{t}, a_{t}^*, s_{t+1})$ \Comment{Record context}
        \EndFor
        \State $R \leftarrow R \cup C$ \Comment{Update the episodic replay buffer}
        \State Update $g$ by maximizing the ELBO in Eq.~(\ref{equ:elbo}) with $R$ \Comment{Model learning}
    \EndFor
  \end{algorithmic}
\end{algorithm*}

In this paper, we propose to solve Eq.~(\ref{equ:opt}) with a sampling-based model-predictive control (MPC) approach. We use MPC for its implementation simplicity, time flexibility, and risk aversion. Also, this sampling-based method makes no assumptions on the pattern of state distributions. Denoting the planning horizon with $\tau_p$, we first define the augmented objective function for an action sequence $A = a_{t_0:t_0 + \tau_p}$ as:
\begin{equation}
\label{equ:ra}
\begin{split}
      \bar{R}(A) \coloneqq & \sum_{t=t_0}^{t_0+\tau_p} [ r\left(s_t, a_t\right) - \lambda[(\mathds{1}(\Pr(s_t \notin \mathcal{S}_{safe}) > \delta) \\
     &+\mathds{1}(\Pr(\hat{s}_t \notin \mathcal{S}_{safe}) > \delta) + \mathds{1}(a_t\notin\mathcal{A}_{safe}) ) ]]
\end{split}
\end{equation}
where $\mathds{1}(Z)$ is the indicator function that returns 1 if $Z$ is true, otherwise 0. 
$s_t$ and $\hat{s}_t$ are the state particles defined in Eq.~(\ref{equ:opt}) and are produced by the trajectory sampling procedure where the uncertainties are propagated (Alg~\ref{alg:traj}).
$\lambda$ serves as the Lagrangian multiplier of the dual problem of Eq.~(\ref{equ:opt}). 
In this paper, we regard $\lambda$ as a fixed hyperparameter and make it sufficiently large 
\begin{equation}
    \lambda \geq \max(|r|)*\tau
\end{equation}
so that the augmented performance is monotonically decreasing w.r.t. the safety violation number.
Considering the uncertainty in the probabilistic model, we evaluate $A$ with the \textit{Conditional Value at Risk (CVaR)}~\cite{tamar2015optimizing} of $\bar{R}(A)$ to make the solutions risk-averse:
\begin{equation}
    \text{CVaR}_\alpha(\bar{R}(A)) = \mathbb{E}\left[\bar{R}(A)|\bar{R}(A) \leq \nu_\alpha(\bar{R}(A))\right]
\end{equation}
where $\alpha \in (0,1)$ and $\nu_\alpha$ is the $\alpha$-quantile of the distribution of $\bar{R}(A)$. In other words, we prefer action sequences with higher CVaR. We then take the first action in the most preferred action sequence and execute it. Instead of uniformly sampling $A$ every time, we utilize the \textit{Cross-Entropy Method (CEM)} as suggested in~\cite{chua2018deep} to keep the historical information. The complete algorithm along with the model-learning part is shown in Alg.~\ref{alg:sa}.

\section{Experiment}
\label{sec:6}
For the evaluation of the proposed algorithm, we aim to answer the following questions through empirical experiments: can CASRL 1) adapt faster to unseen environments with a stream of non-stationary data than existing approaches? 2) reduce the safety violation rate with prior safety constraints?  3) scale to high-dimensional tasks?

\begin{figure}[t]
\centering
    \includegraphics[width=0.98\columnwidth]{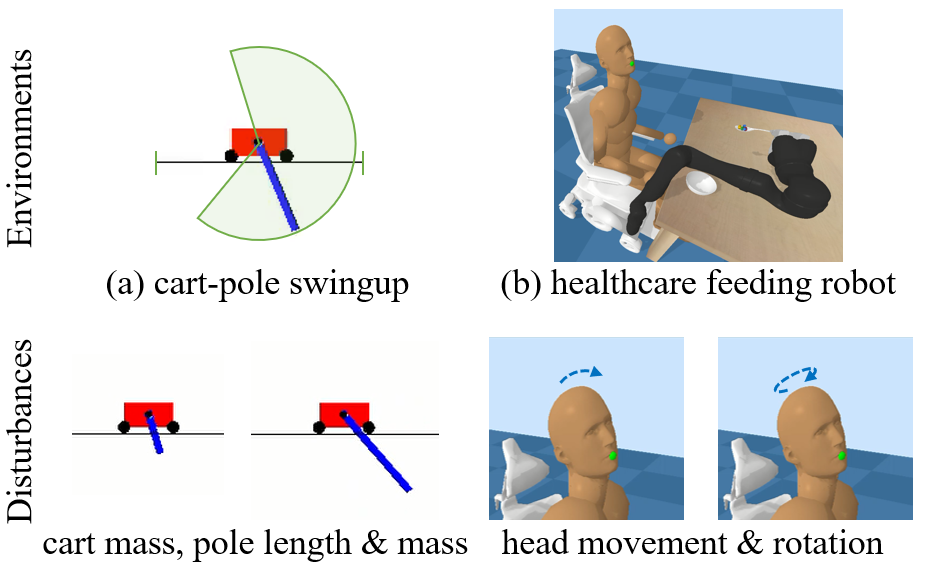} 
\caption{Tasks with non-stationary disturbances and safety constraints.}
\label{fig:envs}
\end{figure}
\subsection{Environments}
To answer the above questions, we test CASRL in two continuously-parameterized non-stationary environments with safety constraints. The setup of the environments (Fig.~\ref{fig:envs}) is introduced below.




\begin{itemize}
    \item \textbf{cart-pole.} ($\mathcal{S} \subseteq \mathbb{R}^4$, $\mathcal{A} \subseteq \mathbb{R}^1$) This is the cart-pole swingup experiment proposed in~\cite{saemundsson2018meta}. The goal is to swing the pole upright by applying force on the cart while keeping the cart close to the center of the rail. We add constraints on the pole angle $\theta \in [-10\degree, 225\degree]$ so that the pole should be swung up from the right side without too much overshoot. We make the task non-stationary by changing the pole length $l$, the pole mass $p_m$, and the cart mass $c_m$ at the beginning of each episode. The observation includes the position $x$ and velocity $\Dot{x}$ of the cart, as well as the angle $\theta$ and angular velocity $\Dot{\theta}$ of the pole. The reward function is $r = \exp\left(-\frac{(x-l\sin{\theta})^2 + (l-l\cos{\theta})^2}{l^2}\right)$ and the highest reward $r=1$ is acquired when the cart is at the center of the rail ($x=0$) and the pole is upright ($\theta=0$). The simulation frequency is 20 Hz.
    
    \item \textbf{healthcare feeding robot.} ($\mathcal{S} \subseteq \mathbb{R}^{23}$, $\mathcal{A} \subseteq \mathbb{R}^7$) The environment is provided by~\cite{erickson2020assistivegym}. The goal is to deliver the medicines to the patient's mouth with a control arm. To keep safe, there should be no direct contact between the patient and the robot. 
    In each episode, the patient moves forward and rotates his head in 4 degree-of-freedom with randomly sampled speeds ($a_f, a_\theta, a_\phi, a_\psi$), which is the disturbance we designed to simulate different preferences. This is a relatively high-dimensional environment and is used to test the scalability of the algorithms. The observation includes the position of the robot joints and the spoon, as well as the position and orientation of the human head.
    The reward function has three parts: $r = r_{dis} + r_{med} + r_{act}$, where $r_{dis}$ penalizes the distance between the spoon and the target position, $r_{med}$ is a large positive value if medicine particles are successfully delivered or a large negative value if they are spilled, and $r_{act}$ penalizes the magnitude of the control input. The simulation frequency is 10 Hz.
\end{itemize}


\subsection{Baselines}
We compare our method with the following baselines:

\begin{itemize}
    \item \textbf{Projection-Based Constrained Policy Optimization (PCPO):} A projection-based safe RL algorithm~\cite{yang2020projection}. The learned policy is projected to the safe region during training.
    \item \textbf{Probabilistic Ensemble and Trajectory Sampling (PETS):} To evaluate the importance of context-aware adaptation, we compare to PETS~\cite{chua2018deep}, a state-of-the-art model-based RL approach.
    \item \textbf{Model-Agnostic Meta-Learning (MAML):} We use the gradient-based MAML~\cite{finn2017model,nagabandi2018learning} to learn the dynamics of the non-stationary environments. The dynamics model is represented by a neural network which is initialized from a pre-trained meta-model and updated online with the nearest context data. \footnote{We used a publicly available implementation at \url{https://github.com/iclavera/learning_to_adapt}.}
    \item \textbf{CASRL without prior safety constraint:} To show whether the prior safety constraint can effectively reduce the safety violation rate, we add another baseline that follows the same structure of CASRL but does not apply the prior safety constraint.
\end{itemize}

Each algorithm (including the proposed method) is first pre-trained in non-safety-critical simulators without any disturbances ($\mathcal{T}_{pre}$) to learn the prior model $h$, where the safety constraints are not applied so that we have enough data from both safe and unsafe regions. We then use these initialized models to safely adapt in disturbance spaces $\mathcal{T}_{adapt}$ to learn the disturbance model $g$, with constraints applied. As introduced in Sec.~\ref{sec:3}, we re-sample the parameters of the environments from $\mathcal{T}_{adapt}$ at the beginning of each episode. The results will reflect the adaptability of the tested algorithms. $\mathcal{T}_{pre}$ and $\mathcal{T}_{adapt}$ used in the experiments are shown in Table~\ref{tab:t_range}.

\begin{table}[h]
\small
\centering
\caption{Disturbance Space. $\mathcal{U}(\cdot)$ denotes uniform distribution.}\smallskip
\label{tab:t_range}
\begin{tabular}{|c |c |c| c|} 
 \hline
 Environment & $\mathcal{T}_{pre}$ & $\mathcal{T}_{adapt}$ & Unit\\
 \hline
 \multirow{3}{*}{cart-pole} & $l = 0.6$  & $l \sim \mathcal{U}[0.2, 1.0]$   & m\\ 
 & $p_m = 0.6$  & $p_m \sim \mathcal{U}[0.2, 1.0]$ & kg\\ 
  & $c_m = 0.6$  & $c_m \sim \mathcal{U}[0.2, 1.0]$ & kg\\ 
 \hline
 \multirow{4}{*}{healthcare} & $a_{f} = 0$  &$a_{f} \sim \mathcal{U}[-1.0, 1.0]$  &\degree/s  \\ 
 & $a_{\theta} = 0$  & $a_{\theta} \sim \mathcal{U}[-2.0, 2.0]$ &\degree/s\\ 
  & $a_{\phi} = 0$  & $a_{\phi} \sim \mathcal{U}[-2.0, 2.0]$ &\degree/s\\ 
   & $a_{\psi}= 0$  & $a_{\psi} \sim \mathcal{U}[-2.0, 2.0]$ &\degree/s\\ 
 \hline
\end{tabular}
\end{table}

In the implementation, we use a hidden size of $[128, 128]$ for all MLP networks. The latent dimension is 8 for the deterministic encoder and latent encoder in the ANP model for both experiments. The planning horizon $\tau$ is set to be 20. Each experiment was run with 10 random seeds. We make the controller risk-averse by setting $\delta=0$ in Equ.~\ref{equ:ra}. All hyperparameters are fine-tuned manually and are provided in our submitted code base. 


\subsection{Result Analysis}

\subsubsection{During Adaptive Training}
\label{sec:during}
\begin{figure}[t]
    \centering
    \includegraphics[width=\columnwidth]{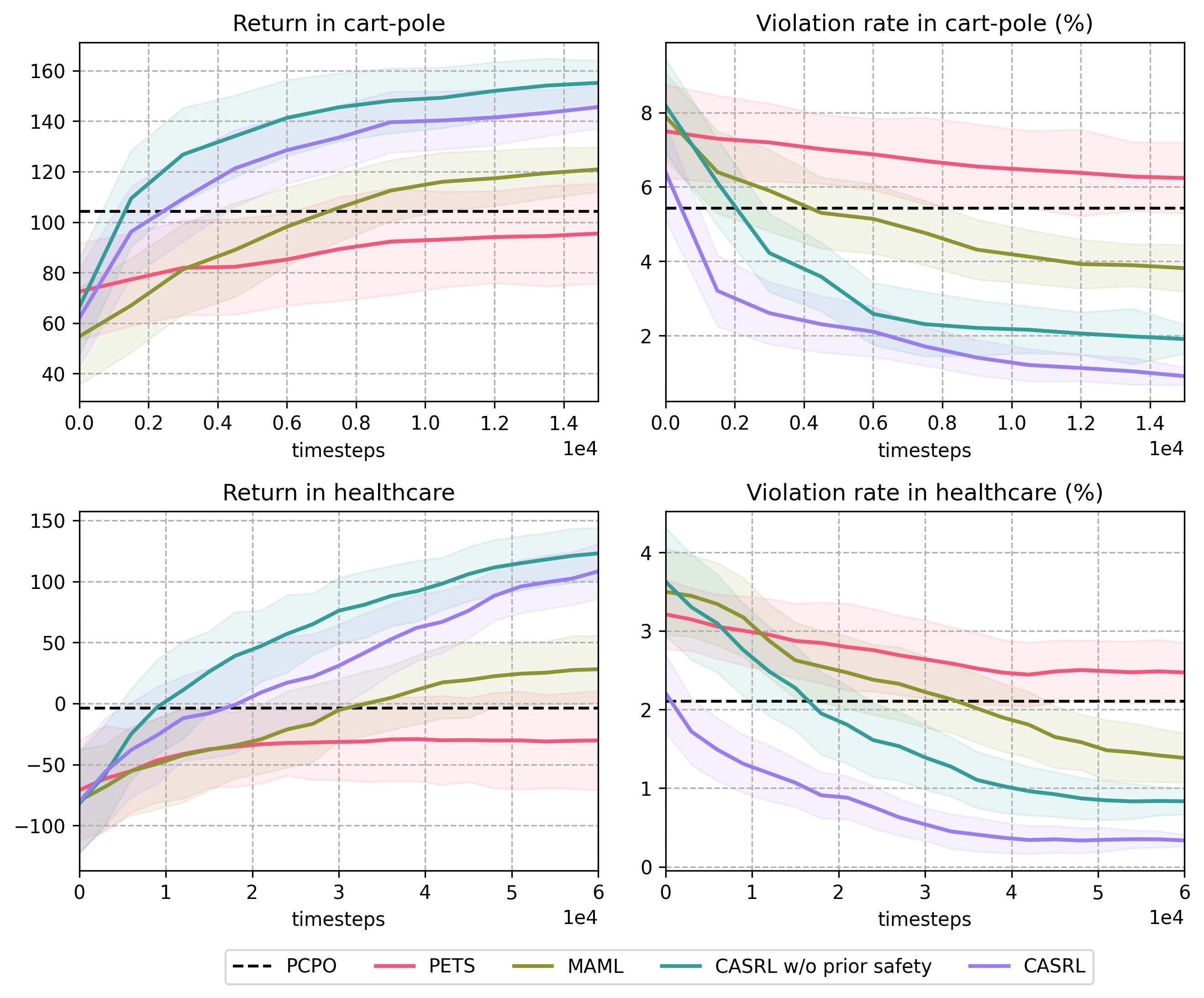} 
    \caption{Return and violation rate during adaptive training. The proposed method CASRL greatly reduces safety violation rate while outperforming MAML in average return.}
    \label{fig:training}
\end{figure}

The average returns and safety violation rates during adaptive training are shown in Fig.~\ref{fig:training}. The violation rate represents the proportion of safety violation time steps in the whole episode. For PCPO, we only plot the highest average performance after its convergence since it requires a lot more samples to train than other model-based methods.
It is shown that the performance of PCPO is limited since it cannot deal with non-stationary environment disturbances.
Though PETS outperforms other methods in most environments during the early stage of training, it fails to continue improving due to the lack of adaptability in non-stationary environments. 
The proposed approach, CASRL, outperforms MAML in both average returns and safety violation rates, especially in the healthcare environment. 
There are two possible reasons. One is that the adaptation of MAML relies on online training of a high-dimensional neural-network model in each step, which is very sensitive to the learning rate and could be unstable in high-dimensional spaces. On the other hand, CASRL only performs online inference. The other possible reason is that MAML cannot model the uncertainties in the environment, which is accomplished by CASRL with a probabilistic latent variable model.

We can also observe that the prior safety constraint can significantly reduce the violation rate with minimal performance degradation.

\begin{figure*}[t]
\begin{subfigure}{.5\textwidth}
  \centering
  \includegraphics[width=.999\linewidth]{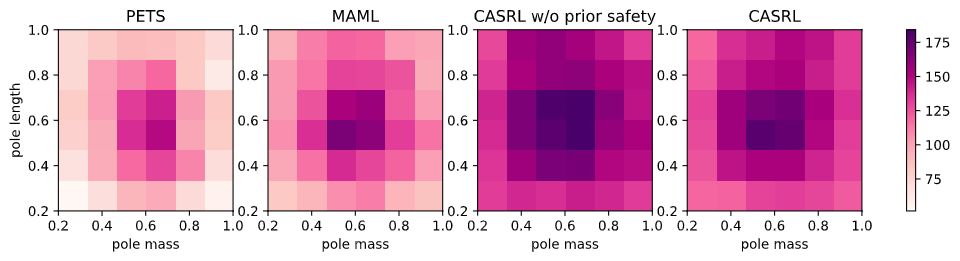}  
  \caption{Return in cart-pole}
  \label{fig:ret_cp}
\end{subfigure}
\begin{subfigure}{.5\textwidth}
  \centering
  \includegraphics[width=.999\linewidth]{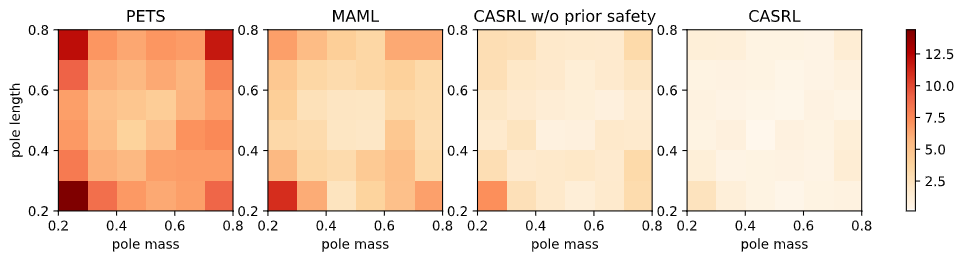}  
  \caption{Violation rate (\%) in cart-pole}
  \label{fig:vr_cp}
\end{subfigure}
\begin{subfigure}{.5\textwidth}
  \centering
  \includegraphics[width=.999\linewidth]{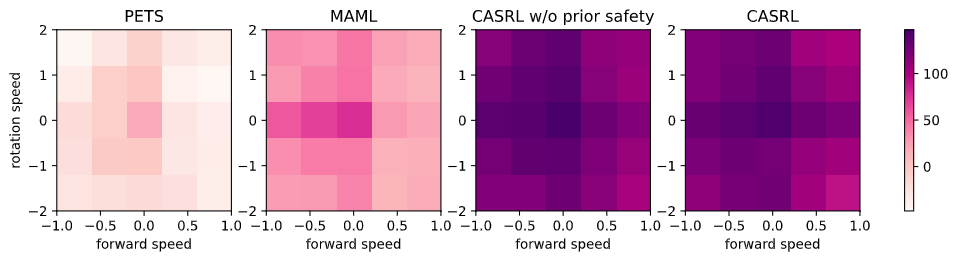}  
  \caption{Return in healthcare}
  \label{fig:ret_heal}
\end{subfigure}
\begin{subfigure}{.5\textwidth}
  \centering
  \includegraphics[width=.999\linewidth]{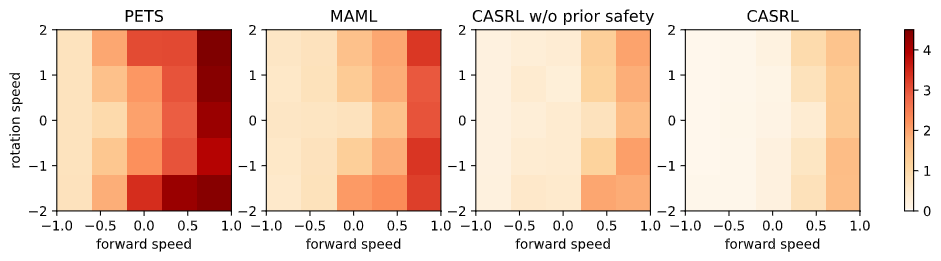}  
  \caption{Violation rate (\%) in healthcare}
  \label{fig:vr_heal}
\end{subfigure}
\caption{Return and violation rate after adaptive training in cart-pole and healthcare environments.}
\label{fig:adaptation}
\end{figure*}

\subsubsection{After Adaptive Training}
We evaluate the performance of models after adaptive training by experiment in the whole disturbance space $\mathcal{T}_{adapt}$ (Tab.~\ref{tab:t_range}). The results of average returns and safety violation rates in cartpole-swingup and healthcare are shown as heatmaps in Fig.~\ref{fig:adaptation}. It is interesting to observe that different constraint functions can lead to different patterns of heatmaps. In the cartpole-swingup environment, most constraint-violation cases concentrate at the corners of the disturbance space (Fig.~\ref{fig:vr_cp}) because the dynamics models in the corners are the most different from the center. In the healthcare environment, however, most constraint-violation cases take place when the human head has a high velocity of forward movement (Fig.~\ref{fig:vr_heal}), which is reasonable since forward movement decreases the distance between the human head and the robot, increasing the risk of direct contact.
Among the methods tested, CASRL shows great robustness and adaptability to disturbances compared to other baselines.

\begin{figure}[t]
    \centering
    \includegraphics[width=\columnwidth]{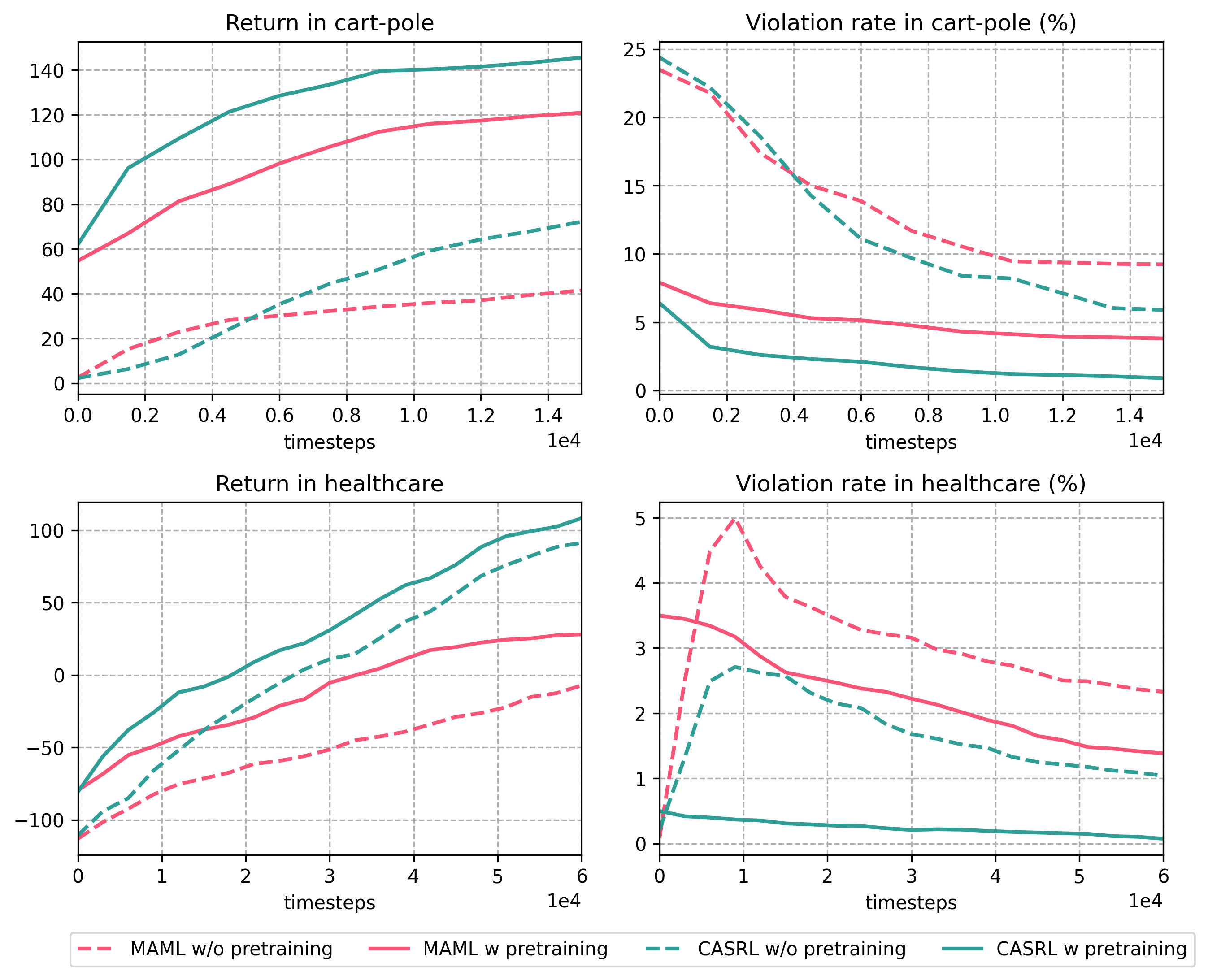} 
    \caption{Comparison of CASRL and MAML with and without the pre-training phase.}
    \label{fig:pre}
\end{figure}



\subsubsection{Effect of pre-training}
\label{sec:pre}
The pre-training phase is essential for CASRL. The pre-trained prior model $h$ not only provides a start point for adaptive learning but also forms the prior safety constraint that improves the safety of the learning process. To show this, we compare the performance of CASRL with and without pre-training in Fig.~\ref{fig:pre}. MAML provides a baseline. It is clearly shown that the pre-training phase significantly benefits the learning process, especially for CASRL.

For the healthcare experiment, the violation rate experienced a big jump in the early stage of training for both methods. The reason is that the robot needs to learn to control its arm before it can approach the patient and possibly violate the safety constraint.


\subsubsection{Effect of prioritized sampling}
\label{sec:pri}
\begin{figure}[h]
\centering
\includegraphics[width=0.95\columnwidth]{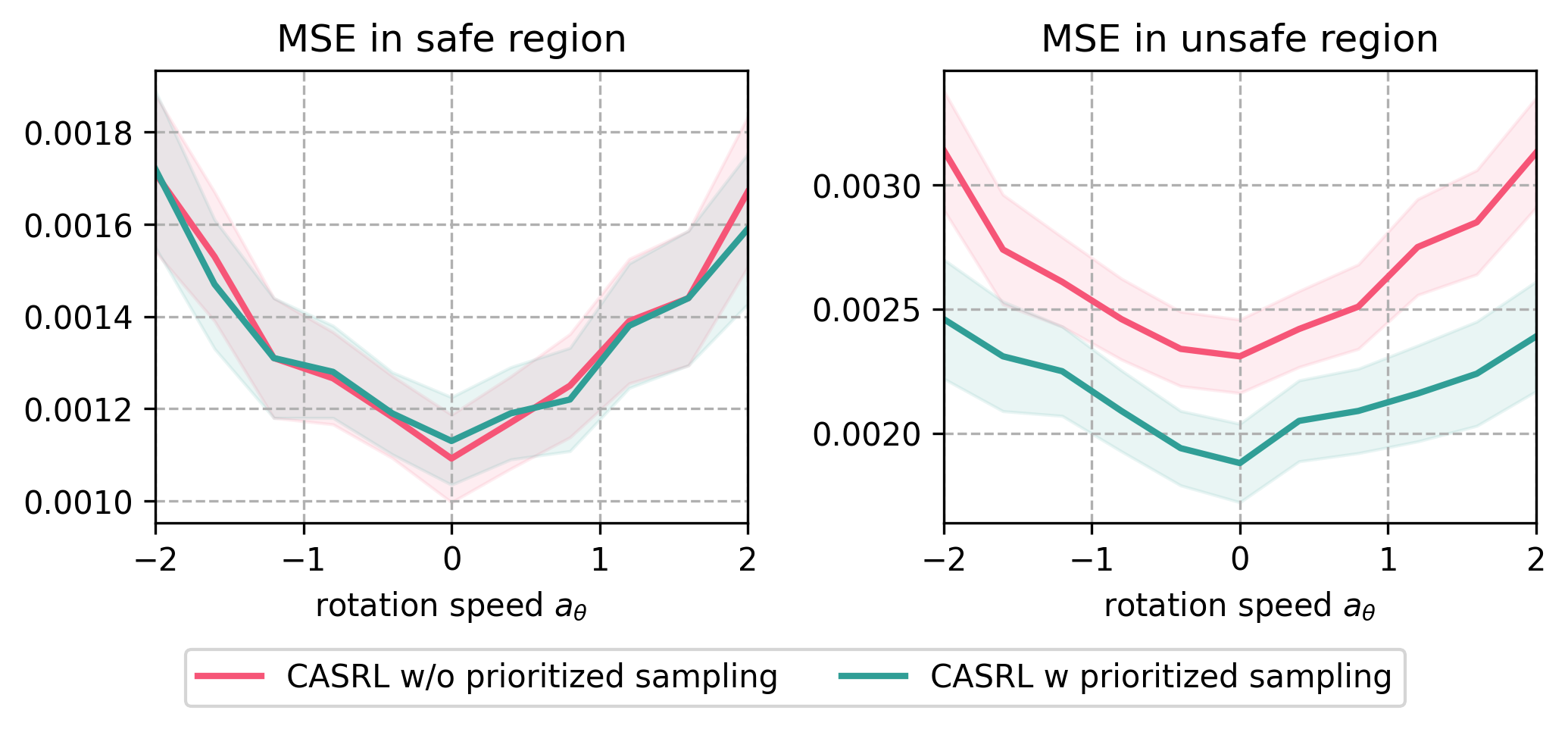} 
\caption{The MSE of single-step dynamics predictions by CASRL in healthcare environment. The prediction accuracy in the unsafe region is improved by prioritized sampling.}
\label{fig:mse}
\end{figure}

We evaluate the effectiveness of prioritized sampling by comparing the mean square error (MSE) of dynamics predictions in safe and unsafe regions. The results are shown in Fig.~\ref{fig:mse}. The prediction accuracy in the unsafe state region is improved by prioritized sampling, while the performance in the safe state region is not influenced. 
The reason could be that without prioritized sampling, the model is biased towards the safe data due to the rareness of the unsafe samples.

\section{Conclusion}

In this paper, we propose the context-aware safe reinforcement learning (CASRL) method as a meta-learning framework to realize safe adaptation in non-stationary environments. The non-stationary disturbances are identified with a probabilistic latent variable model by online Bayesian inference.
A risk-averse model-predictive controller is used for safe planning with uncertainties, where we incorporate prior safety constraints to enable fast adaptation with prior knowledge.
We also utilize prioritized sampling of unsafe data to alleviate the data imbalance in safety-critical environments.
The algorithm is evaluated in both toy and realistic high-dimensional environments. Results show that CASRL significantly outperforms existing baselines in terms of safety and robustness.

Although CASRL is potentially beneficial for RL applications in safety-critical tasks, it may have its limitations. 
For example, the disturbance space could be much larger if we use image inputs with noises. Although the ANP model has been shown to work for image reconstruction tasks~\cite{kim2019attentive}, it may fail for dynamics prediction in complex environments. In that case, one potential solution is to conduct dynamics prediction in the latent space as in Dreamer~\cite{hafner2019dream}, which is directly applicable for CASRL. The hyperparameter-tuning for learning rates, network structures, and especially the latent dimensions could be another challenge for CASRL.








\bibliographystyle{IEEEtran}
\bibliography{main}






\end{document}